\pdfoutput=1

\documentclass[11pt]{article}
\usepackage{amsmath} 
\usepackage[preprint]{acl}
\usepackage{natbib}

\usepackage{times}
\usepackage{latexsym}

\usepackage[T1]{fontenc}

\usepackage[utf8]{inputenc}

\usepackage{microtype}
\usepackage{booktabs}

\usepackage{inconsolata}

\usepackage{graphicx}
\usepackage{hyperref}
\usepackage{amssymb}
\usepackage{float}
\usepackage{multirow} 
\usepackage{pifont}       
\usepackage{bbding}
\usepackage{fontawesome}
\usepackage{inconsolata}
\usepackage{makecell}
\usepackage{listings}
\usepackage{subcaption}
\usepackage{CJKutf8}
\graphicspath{{./img/}}
\title{Know-MRI: A Knowledge Mechanisms Revealer\&Interpreter for Large Language Models}

\author{
    \textbf{Jiaxiang Liu}\textsuperscript{$*$ 1,2}\textbf{,} \textbf{Boxuan Xing}\textsuperscript{$*$ 2}\textbf{,} \textbf{Chenhao Yuan}\textsuperscript{$*$ 2}\textbf{,} \textbf{Chenxiang Zhang}\textsuperscript{1}\textbf{,} \textbf{Di Wu}\textsuperscript{1}\textbf{,} \\ \textbf{Xiusheng Huang}\textsuperscript{1,2}\textbf{,} \textbf{Haida Yu}\textsuperscript{1,2}\textbf{,} \textbf{Chuhan Lang}\textsuperscript{2}\textbf{,} \textbf{Pengfei Cao}\textsuperscript{$\dag$ 1,2}\textbf{,} \textbf{Jun Zhao}\textsuperscript{1,2}\textbf{,} \textbf{Kang Liu}\textsuperscript{$\dag$ 1,2,3}
    \\
    \textsuperscript{1}The Key Laboratory of Cognition and Decision Intelligence for Complex Systems, \\Institute of Automation, Chinese Academy of Sciences, Beijing, China\\
    \textsuperscript{2}School of Artificial Intelligence, University of Chinese Academy of Sciences \\
    \textsuperscript{3}Shanghai Artificial Intelligence Laboratory\\
    liujiaxiang21@mails.ucas.ac.cn, \{pengfei.cao, jzhao, kliu\}@nlpr.ia.ac.cn
}

\begin{document}
\maketitle
\renewcommand{\thefootnote}{\fnsymbol{footnote}}
\footnotetext[1]{Equal contribution.}
\footnotetext[2]{Corresponding authors.}
\renewcommand{\thefootnote}{\arabic{footnote}}
\begin{abstract}
As large language models (LLMs) continue to advance, there is a growing urgency to enhance the interpretability of their internal knowledge mechanisms. Consequently, many interpretation methods have emerged, aiming to unravel the knowledge mechanisms of LLMs from various perspectives. However, current interpretation methods differ in input data formats and interpreting outputs. The tools integrating these methods are only capable of supporting tasks with specific inputs, significantly constraining their practical applications. To address these challenges, we present an open-source \textbf{Know}ledge \textbf{M}echanisms \textbf{R}evealer\&\textbf{I}nterpreter (\textbf{Know-MRI}) designed to analyze the knowledge mechanisms within LLMs systematically. Specifically, we have developed an extensible core module that can automatically match different input data with interpretation methods and consolidate the interpreting outputs. It enables users to freely choose appropriate interpretation methods based on the inputs, making it easier to comprehensively diagnose the model's internal knowledge mechanisms from multiple perspectives. Our code is available at \href{https://github.com/nlpkeg/Know-MRI}{https://github.com/nlpkeg/Know-MRI}. We also provide a demonstration video on \href{https://youtu.be/NVWZABJ43Bs}{https://youtu.be/NVWZABJ43Bs}. 
\end{abstract}

\section{Introduction}

\begin{table*}
\centering \scalebox{0.85}{
\begin{tabular}{c c c c c c c}\toprule[2pt] \hline
			\multirow{3}{*}{Toolkit}& \multicolumn{6}{c}{Feature} \\ \cmidrule(lr){2-7}
            &\multirow{2}{*}{Input format}&\multicolumn{2}{c}{Perspective}&\multirow{2}{*}{Flexibility}&\multirow{2}{*}{Extensibility}&\multirow{2}{*}{User-friendly}  \\ \cmidrule(lr){3-4}
            &&Internal&External&&\\ \hline
            LIT&\textcolor{orange}{Fair}&\thead{\textcolor{orange}{\texttt{Embedding}, \texttt{Attention}}}&\thead{\textcolor{red}{None}}&\textcolor{orange}{Fair}&\textcolor{red}{\ding{54}}&\textcolor{green}{Good}\\
            
            Ecco&\textcolor{orange}{Fair}&\thead{\textcolor{red}{None}}&\thead{\textcolor{orange}{\texttt{Attribution}}}&\textcolor{red}{Poor}&\textcolor{red}{\ding{54}}&\textcolor{orange}{Fair}\\
            
             LM-Debugger&\textcolor{red}{Single}&\thead{\textcolor{orange}{\texttt{MLP/Neuron}}}&\thead{\textcolor{red}{None}}&\textcolor{red}{Poor}&\textcolor{red}{\ding{54}}&\textcolor{green}{Good}\\
             
             VISIT&\textcolor{red}{Single}&\thead{\textcolor{orange}{\texttt{Hiddenstate}, \texttt{MLP/Neuron},}\\ \textcolor{orange}{\texttt{Attention}}}&\thead{\textcolor{red}{None}}&\textcolor{red}{Poor}&\textcolor{red}{\ding{54}}&\textcolor{orange}{Fair}\\
             
             Inseq&\textcolor{red}{Single}&\thead{\textcolor{orange}{\texttt{MLP/Neuron}}}&\thead{\textcolor{orange}{\texttt{Attribution}}}&\textcolor{orange}{Fair}&\textcolor{red}{\ding{54}}&\textcolor{orange}{Fair}\\
             
            LM-TT&\textcolor{red}{Single}&\thead{\textcolor{orange}{\texttt{Attention}, \texttt{MLP/Neuron}}}&\thead{\textcolor{red}{None}}&\textcolor{red}{Poor}&\textcolor{red}{\ding{54}}&\textcolor{green}{Good}\\
            
             \textbf{Know-MRI}&\textcolor{green}{Diverse}&\textcolor{green}{All}&\textcolor{green}{All}&\textcolor{green}{Good}&\textcolor{green}{\ding{52}}&\textcolor{green}{Good}\\
            \hline
            \bottomrule[2pt]
		\end{tabular}}
\caption{Comparison of existing interpretation toolkits. Input format refers to the diversity of the input data format. Perspective refers to the interpreting form of the methods (detailed categorization is listed in Section \ref{related}) involved in the toolkit. Flexibility refers to how well the toolkit can select appropriate interpretation methods for specific inputs. Extensibility refers to the capability to accommodate additional interpretation methods. User-friendly refers to the ease of use of the toolkit.}
\label{related-work}
\end{table*}

\begin{figure}[!ht]
	\centering
        \includegraphics[width=0.95\linewidth]{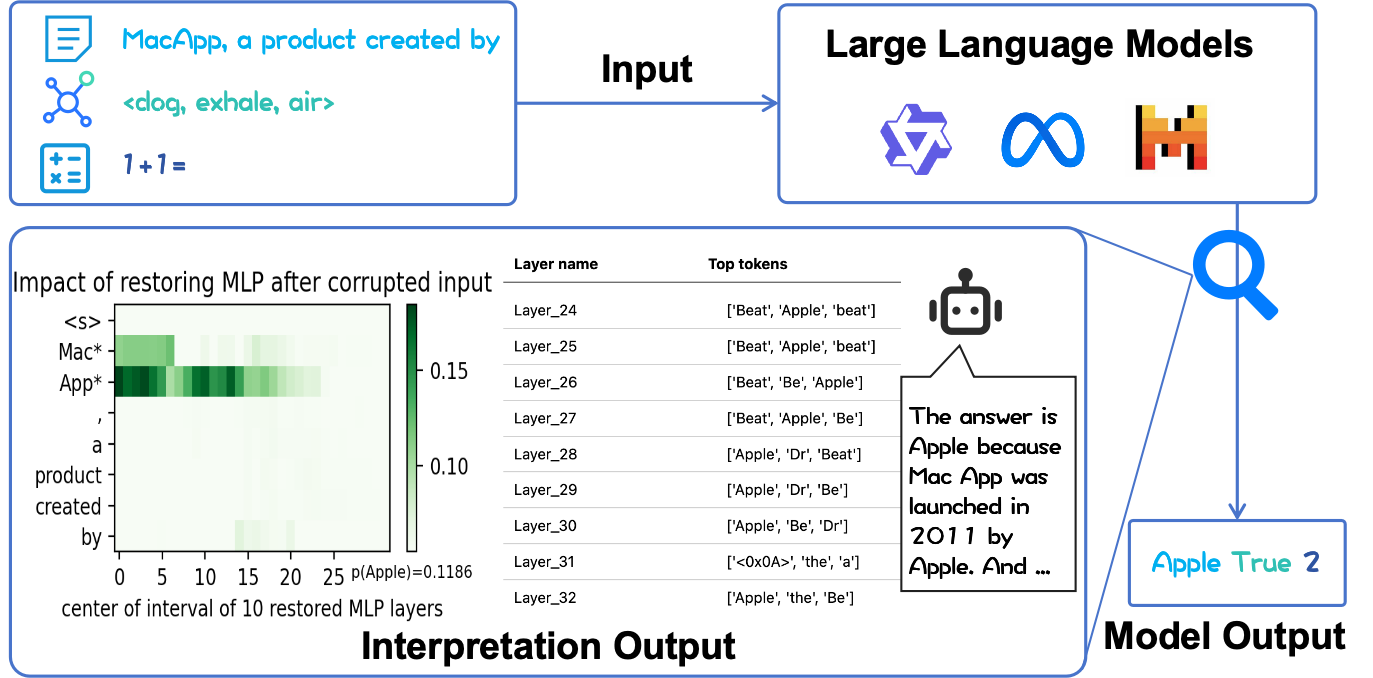}
		\caption{Illustration of LLMs interpretation.}
        \label{intro}
\end{figure}

Large language models (LLMs), accumulating a vast amount of factual knowledge through extensive pre-training corpora, are often seen as parameterized knowledge bases \citep{gpt2, gpt-j, jiang2023mistral7b, llama2, openai2024gpt4technicalreport, qwen2.5, DeepSeek-R1}. However, the underlying knowledge mechanisms of LLMs—including how they learn, store, utilize, and evolve knowledge \citep{knowledge-mechanisms}—remain poorly understood. This lack of transparency poses significant challenges to the safe and trustworthy deployment of LLMs across sensitive domains such as healthcare, finance, and the judiciary. Aiming to reveal the knowledge mechanisms in LLMs, as shown in Figure \ref{intro}, current interpretation methods often generate different kinds of interpretation results (such as figures with tracing weights, unembedding tables, explanation texts) according to the input (such as the targeted knowledge) with different formats (such as textual prompts, triples, mathematical operations) \citep{editingsolution, chen2023journeycenterknowledgeneurons, chen2025knowledgelocalizationmissionaccomplished, Microscop}.

To enhance the community's understanding of the knowledge mechanism of LLMs, a growing number of interpretation tools have been developed \citep{LIT, alammar-2021-ecco, lmdebugger, katz-belinkov-2023-visit, sarti-etal-2023-inseq, lmtt}. Although these tools have propelled interpretation research forward, as summarized in Table \ref{related-work}, they have four interconnected limitations: 1) \textbf{Single Input Format}: Due to the various forms of knowledge, existing tools mainly support \textit{limited input data formats}, such as a single prompt, causing inconvenience to the users' usage. 2) \textbf{Biased Interpretation}: The diversity of interpretation methods causes existing tools to \textit{focus narrowly on specific interpreting perspectives}. 3) \textbf{Low Flexibility and Extensibility}: Existing tools cannot flexibly select interpretation methods based on input. They also exhibit low extensibility on new models, data, and interpretation methods. 4) \textbf{Less User-friendly}: Current toolkits are primarily designed for domain experts, making them \textit{less user-friendly}, particularly for beginners.

To address the aforementioned issue, the paper proposes \textbf{Know-MRI}, a \textbf{Know}ledge \textbf{M}echanisms \textbf{R}evealer\&\textbf{I}nterpreter for LLMs. As shown in Figure \ref{framework}, the characteristic of Know-MRI’s key feature is its ability to select the appropriate interpretation method based on the input data by matching the \texttt{support\_template\_keys} (Dataset) with the \texttt{requires\_input\_keys} (Interpretation Method). Additionally, Know-MRI provides an extensible API that allows users to integrate their own interpretation methods, and a UI demo is offered to further enhance user-friendliness. In general, Know-MRI has the following advantages: 1) \textbf{Rich Input Format Support}: In contrast to previous tools that mainly targeted a specific or a limited kind of input, Know-MRI supports a variety of different data formats. Beyond factual knowledge, it can also adapt to different task datasets (such as mathematical reasoning, sentiment analysis, etc.), totally covering 13 datasets with different input formats. 2) \textbf{Methods Diversity}: Know-MRI analyzes LLMs from both internal and external perspectives. Specifically, it can jointly explore internal reasoning processes and external behavioral attributions, supporting 8 classic interpretation methods. 3) \textbf{Flexibility}: For an input, Know-MRI can automatically match the required interpretation methods. 4) \textbf{Extensibility}: Integrating new methods and models into Know-MRI requires only simple encapsulation, making the addition of
new methods straightforward. 4) \textbf{User-friendly}: Know-MRI is meticulously designed to help users quickly understand existing interpretation methods through its user interface, guidelines, and detailed results descriptions.

Additionally, with the help of this toolkit, we conduct a case study making comparisons between similar methods that jointly confirm the significant role of subject in LLMs' handling of factual knowledge. This further demonstrates the effectiveness of Know-MRI.

\section{Related Work} \label{related}

\subsection{Interpretation Methods} 
As shown in Table \ref{inpretmethod-table}, existing knowledge mechanisms interpretation methods can be mainly divided into the following two categories:

\paragraph{External Interpretation:} \textbf{These methods primarily focus on analyzing the input-output relationships from an external perspective.} A direct approach involves eliciting \texttt{Self-explanations} from LLMs. For instance, \citet{selfexplaintion} propose a method that leverages LLMs to identify the contribution of input words to model predictions. In contrast, \texttt{Attribution} \citep{Integrated-Gradients} utilizes gradients to calculate the contribution, offering a mathematically grounded perspective on output attribution. 

\paragraph{Internal Interpretation:} \textbf{This category delves into the decision processes of LLMs by examining their internal representations and modular operations}. From the representation perspective, researchers analyze features through \texttt{Hidden state} \citep{logit-lens, ghandeharioun2024patchscopes} and \texttt{Space probing} \citep{spine}. The analysis of module further dissects functional components along four axes: 1) \texttt{Embedding} \citep{LIT}, 2) \texttt{Attention} \citep{attentionisall}, 3) \texttt{MLP/Neuron} \citep{ROME, knowledgeneuron, fine}, and 4) \texttt{Circuit} \citep{knowledgecircuit}, collectively revealing the architectural foundations of model behavior. The Interpretation Datasets are listed in the Appendix \ref{datasetlisted}.
\begin{figure*}
        \centering
        \includegraphics[width=0.95\textwidth]{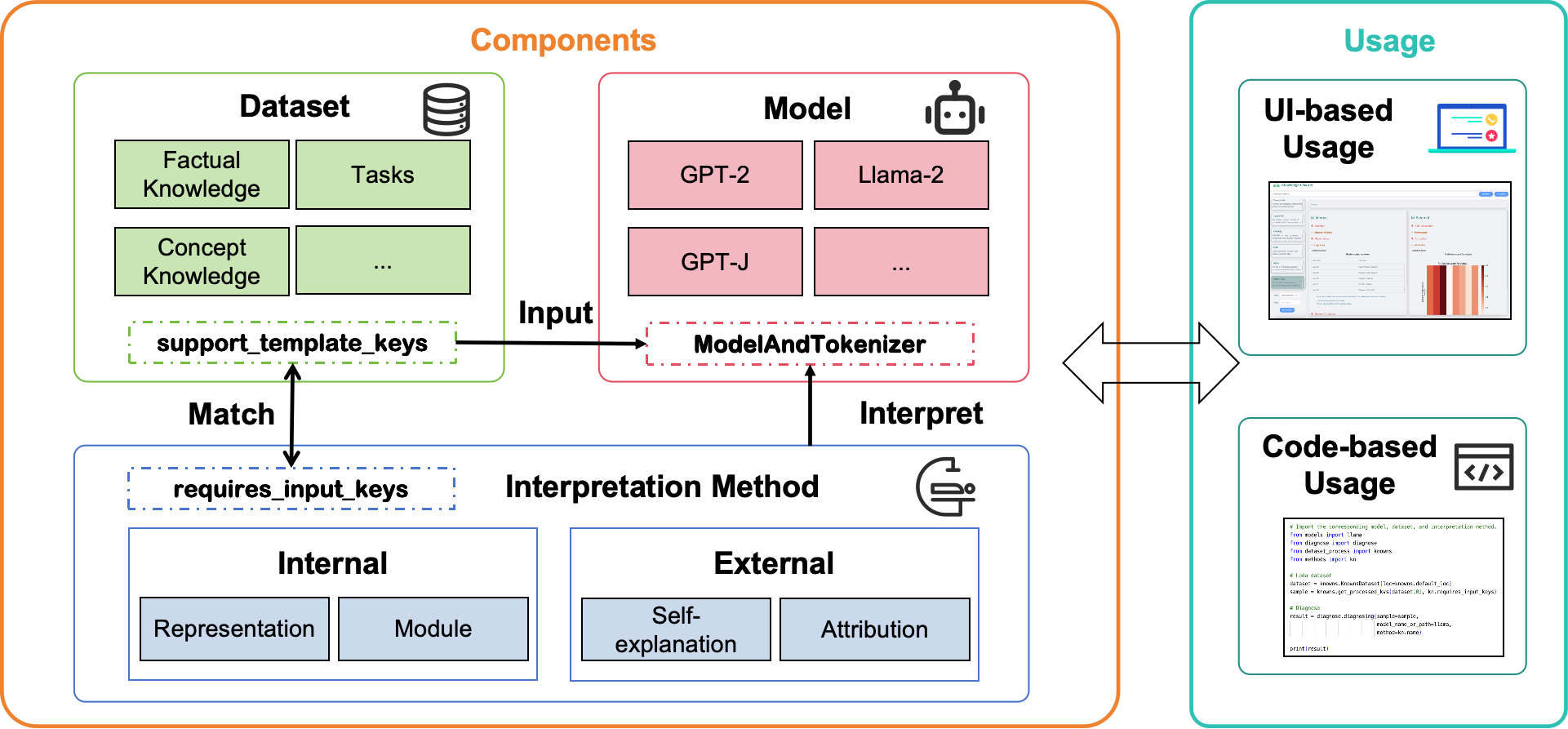}
        \caption{The frame work of Know-MRI. Know-MRI primarily consists of three components: \texttt{Model}, \texttt{Dataset}, and \texttt{Interpretation Method}. Know-MRI can be invoked through either UI or Code. The UI-based usage is designed to assist users in quick learning and utilization. The Code-based usage, on the other hand, has greater extensibility.}
        \label{framework}
\end{figure*}
\subsection{Interpretation Toolkits} 
Recent years have witnessed several interpretation toolkits aimed at enhancing community understanding of LLMs' knowledge mechanisms \citep{LIT, alammar-2021-ecco, lmdebugger, katz-belinkov-2023-visit, sarti-etal-2023-inseq, lmtt}. However, existing methods have differences in their
required input and interpretation output, making it difficult to use these methods in a single toolkit. For instance, the Knowledge Neuron (KN) method \citep{knowledgeneuron} necessitates annotated input data with ground truth and generates corresponding figures for knowledge attribution. Conversely, Patchscopes \citep{ghandeharioun2024patchscopes} works without ground truth but mandates structured tabular for interpretation. Such divergent specifications confine existing toolkits to a few interpretation perspectives or limited input formats, as shown in the ``Perspective'' and ``Input data'' columns of Table \ref{related-work}. Even the relatively generic Inseq \cite{sarti-etal-2023-inseq} cannot flexibly match every input with the interpretation methods and consolidate the outputs. To address the aforementioned issue, we propose a framework capable of automatically pairing inputs with interpretation methods.

\section{Know-MRI Toolkit}
\textbf{Know}ledge \textbf{M}echanisms \textbf{R}evealer\&\textbf{I}nterpreter (\textbf{Know-MRI}) is a unified framework designed to systematically integrate existing interpretation methods, enabling comprehensive analysis of LLMs' knowledge mechanisms. As shown in Figure \ref{framework}, Know-MRI primarily integrates model, dataset, and interpretation method. For a given input and model, Know-MRI can automatically select the corresponding interpretation methods and generate interpreting results. Additionally, Know-MRI also offers UI-based and Code-based usage. In the following section, we will introduce the components of Know-MRI and present the toolkit usage.

\subsection{Toolkit Components}
As outlined above, Know-MRI seamlessly integrates three core components: model, dataset, and interpretation methods. Our exposition of these elements will be structured around two key dimensions: \textit{supported types and extensibility}.

\subsubsection{Model}
\paragraph{Supported Types} Know-MRI can apply to 9 architectures of models on Huggingface\footnote{https://huggingface.co}, including \texttt{Bert} \citep{Bert}, \texttt{GPT2} \citep{gpt2}, \texttt{GPT-J} \citep{gpt-j}, \texttt{T5} \citep{T5}, \texttt{Llama2} \citep{llama2}, \texttt{Baichuan} \citep{baichuan2023baichuan2}, \texttt{Qwen} \cite{qwen2.5}, \texttt{ChatGLM} \citep{glm2024chatglm} and \texttt{InternLM} \citep{InternLM}.

\paragraph{Extensibility} Building upon the architectural insights from \citet{ROME}, we propose a standardized encapsulation approach through the \texttt{ModelAndTokenizer} class. This abstraction layer systematically unifies model interfaces while preserving their intrinsic computational characteristics. To ensure adaptability in the rapidly evolving model ecosystem, Know-MRI allows us to incorporate new types of LLMs. We will implement continuous maintenance for the \texttt{ModelAndTokenizer} class. 

\subsubsection{Dataset} \label{datasetttt}
\paragraph{Supported Types} Know-MRI has integrated more than 13 datasets with different input formats. 

These datasets embrace a rather broad scope. Some involve structured-input, such as ZsRE \citep{zsre}, PEP3k \citep{pep3k} and Know-1000 \citep{ROME}, while others are derived from direct prompts, such as GSM8K \citep{cobbe2021gsm8k}, Imdb \citep{imdb} and Opus 100 \citep{opus_100}. More details are listed in Appendix \ref{dataset}.

\paragraph{Extensibility} Users can incorporate their own datasets by simply integrating the \texttt{Dataset} class in Pytorch\footnote{https://pytorch.org}. It is noteworthy that to facilitate the matching of the corresponding interpretation methods, users need to add the field named \texttt{support\_template\_keys} to indicate which keys the current dataset supports. Specifically, \texttt{support\_template\_keys} is a list that describes the format of inputs included in the current dataset, such as prompt, subject, and ground truth, etc. The introduction about keys is in Appendix \ref{template_keys}. For instance, Known-1000 \citep{ROME} is a question-answering dataset based on factual triplets, and each question encompasses various forms of expressions. Therefore, its \texttt{support\_template\_keys} should be [``prompt'', ``prompts'', ``ground\_truth'', ``triple\_subject'', ``triple\_relation'', ``triple\_object''].

\subsubsection{Interpretation Method}

\paragraph{Supported Types} In Table \ref{inpretmethod-table}, we show that Know-MRI employs eight distinct types of interpretation methods, culminating in a total of eleven interpretation techniques. These techniques fall into two main categories: \textit{external and internal explanations}.  External methods include Self-explanations \citep{selfexplaintion_new} and Attribution \citep{Integrated-Gradients}. Internal explanations are further divided into Module and Representation approaches. From the perspective of Module, we have integrated: 1) \texttt{Embedding}: Projection \citep{LIT}, 2) \texttt{Attention}: Attention Weights \citep{attentionisall}, 3) \texttt{MLP/Neuron}: KN \citep{knowledgeneuron}, CausalTracing \citep{ROME}, FINE \citep{fine}, 4) \texttt{Circuit}: Knowledge Circuit \citep{knowledgecircuit}. Representation can be categorized into: 1) Hiddenstate: Logit Lens \citep{logit-lens}, PatchScopes \citep{ghandeharioun2024patchscopes}, 2) Space probing: SPINE \citep{spine}.

\begin{table}[!ht]
    \centering\resizebox{\linewidth}{!}{
    \begin{tabular}{ccc}
    \toprule[2pt]\hline 
        \multirow{2}{*}{External}&\multicolumn{2}{c}{Internal}\\ \cmidrule(lr){2-3}
        &Module&Representation \\ \hline
        \thead{\texttt{Self-explanations}, \\ \texttt{Attribution}}&\thead{\texttt{Embedding}, \texttt{Attention}, \\ \texttt{MLP/Neuron}, \texttt{Circuit}}&\thead{\texttt{Hiddenstate}, \\ \texttt{Space probing}} \\
        
        \hline 
        \bottomrule[2pt]
    \end{tabular}  
    }
    \caption{The classification of existing interpretation methods.}
    \label{inpretmethod-table}
\end{table}

\paragraph{Extensibility} Users merely need to encapsulate their interpretation methods into a \texttt{diagnose} function. Corresponding to Dataset, users are required to provide a \texttt{requires\_input\_keys} to describe the necessary input for this method. Corresponding to \texttt{support\_template\_keys} in Section \ref{datasetttt}, \texttt{requires\_input\_keys} is also a list. It is indicative of the input format required by the interpretation method. For instance, the Knowledge Neuron (KN) method \citep{knowledgeneuron} necessitates semantically similar input prompts with ground truth. So its \texttt{requires\_input\_keys} should be [``prompts'', ``ground\_truth''].

\subsection{Toolkit Usage}
Know-MRI offers two operational modes: a user interface (UI) and a code-based usage. The following sections will explain how to use Know-MRI through each mode in turn.

\subsubsection{UI-based Usage}
\begin{figure*}
		\centering
		\includegraphics[width=0.95\textwidth]{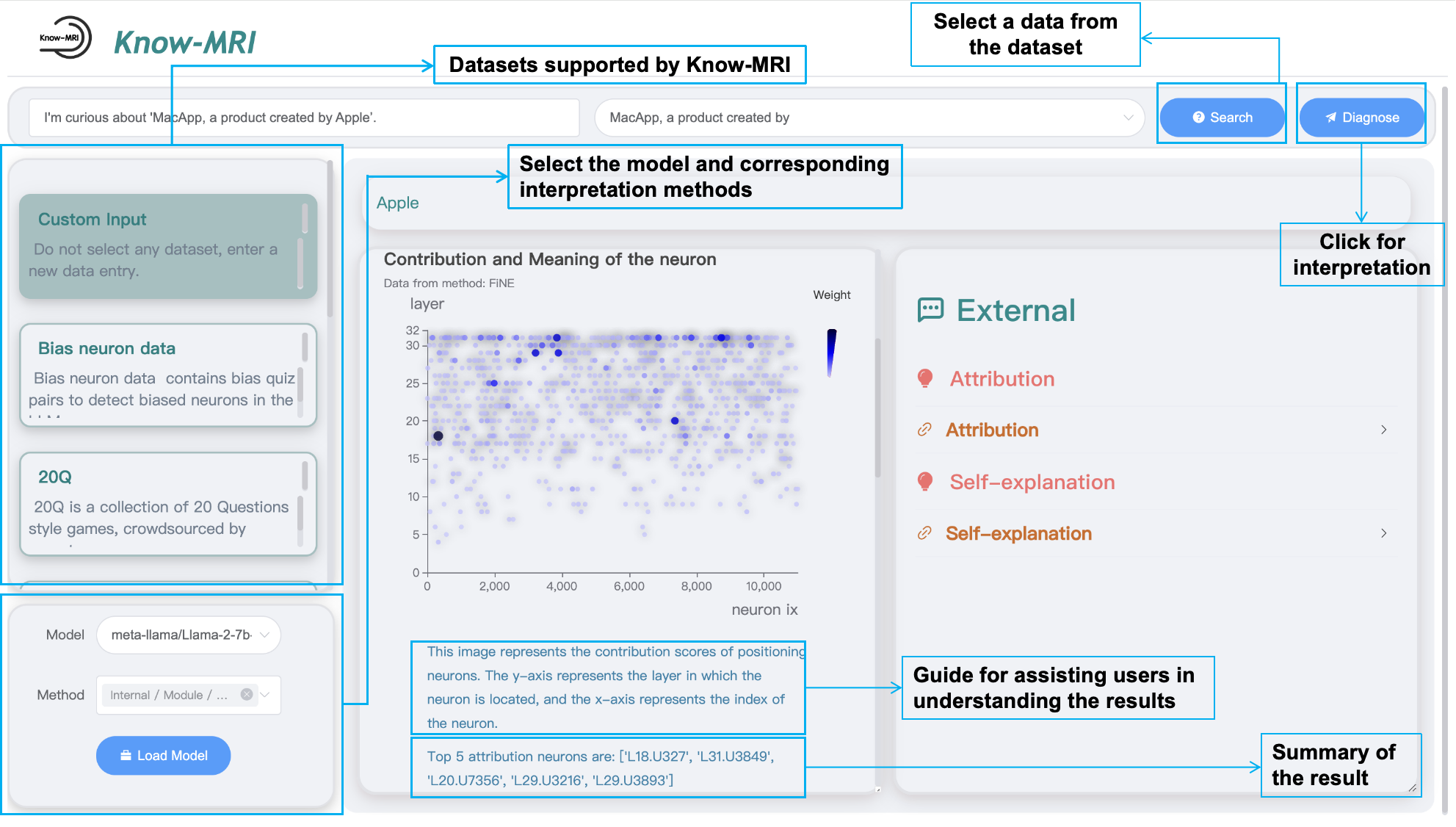}
		\caption{User interface (UI) of Know-MRI.}
        \label{UI}
\end{figure*}

Using a UI-based approach enables beginners to get started more quickly and allows researchers to rapidly invoke existing interpretation methods. As shown in Figure \ref{UI}, Know-MRI's UI is meticulously designed to be intuitive and user-friendly:

\textbf{Know-MRI is easy to use.} Users can comprehensively interpret models with simple click operations. In the upper left corner, users can select their preferred dataset or enter Custom Input. In the lower left corner, they can choose the corresponding model and the interpretation methods provided by Know-MRI. In the top right corner, users can utilize the ``Search'' button to select data and click ``Diagnose'' to perform interpretation. Additionally, Know-MRI integrates several interpretation methods with identical output forms (e.g. KN \citep{knowledgeneuron} and FINE \citep{fine}) to assist users in better comparison.

\textbf{Know-MRI is easy to understand.} For each interpretation method, Know-MRI provides template-based descriptions. As illustrated in Figure \ref{UI}, Know-MRI offers explanations of how to read the results of the KN \citep{knowledgeneuron} and highlights significant points.

\textbf{Know-MRI is flexible in handling user input.} Recognizing that users may occasionally provide imprecise or unconventional queries, Know-MRI employs a dual technique: 1) GPT-4o \citep{gpt4ocard} rewrites users' inputs into the anticipated form. 2) BGE-base \citep{bge_embedding} searches for relevant knowledge within existing datasets. As illustrated in Figure \ref{UI}, Know-MRI effectively handles atypical inputs like \textit{I'm curious about ``MacApp, a product created by Apple''}.

\subsubsection{Code-based Usage}
To enable researchers to efficiently apply existing interpretation methods in experimental settings, Know-MRI implements a code-based usage.

\begin{figure}[!ht]
		\centering
		\includegraphics[width=0.45\textwidth]{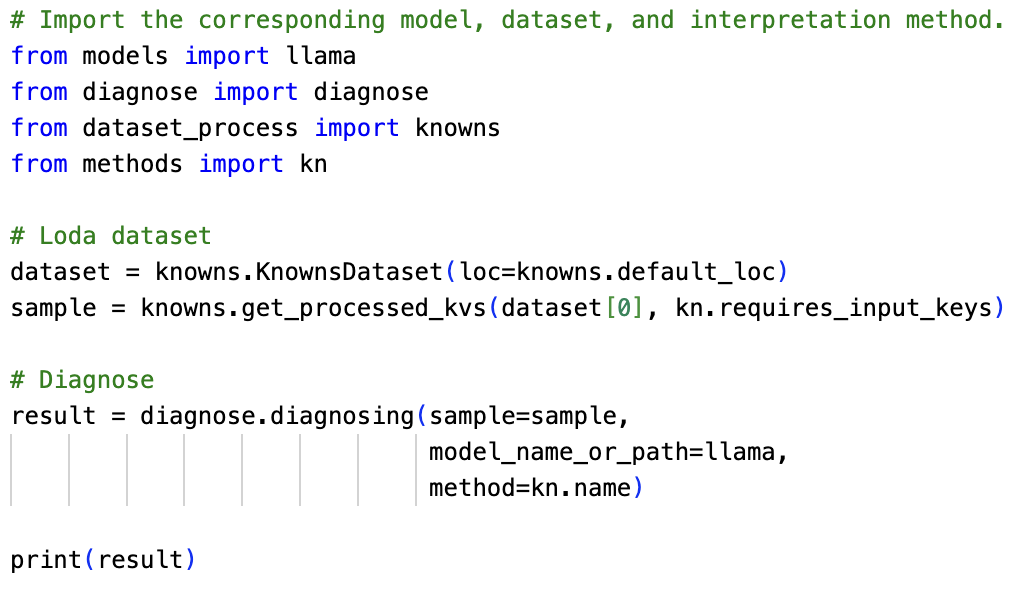}
		\caption{A code example of Know-MRI.}
        \label{code_demo}
\end{figure}

As shown in Figure \ref{code_demo}, the framework demonstrates remarkable operational efficiency by requiring only concise code snippets (8 lines) to implement the KN method \citep{knowledgeneuron} on the dataset Known 1000 \citep{ROME}. The same applies to other interpretation methods as well.

\section{Case Study and Evaluation}
In this section, we will utilize the Know-MRI to evaluate LLMs from three axes: a use case, extended application and human evaluation.

\subsection{Use Case}
In this experiment, we employ the UI-based usage of Know-MRI.
\paragraph{Experimental Setup} Our experiment involves the interpretation of Llama2-7B \citep{llama2} using a random sample from the fundamental knowledge dataset Know 1000.

\paragraph{Result} With the help of Know-MRI, we can have some interesting findings with comparison and thus validate the correctness of Know-MRI.

\begin{CJK}{UTF8}{gbsn}
\begin{table}[!ht]
    \centering\resizebox{\linewidth}{!}{
    \begin{tabular}{ccc}
    \toprule[2pt]\hline 
        Method&Top neurons&Top tokens \\ \hline
        \multirow{4}{*}{FINE}&L18.U327&[``Apple'', ``apple'', ``Mac''] \\
        &L31.U3849&[``Harry'', ``Dick'', ``Frank''] \\
        &L29.U3216&[``Mac'', ``mac'', ``Mac''] \\
        &L29.U3893&[``Apple'', ``Microsoft'', ``Canadian''] \\ \hline
        \multirow{4}{*}{KN}&L1.U6972&[``elin'', ``符'', ``argent''] \\
        &L1.U4503&[``ederb'', ``curity'', ``atos''] \\
        &L29.U3216&[``Mac'', ``mac'', ``Mac''] \\
        &L20.U7356&[``Warner'', ``Sony'', ``companies''] \\
        
        \hline 
        \bottomrule[2pt]
    \end{tabular}  
    }
    \caption{Comparison between top-4 neurons selected by different methods.}
    \label{KN&FINE}
\end{table}
\end{CJK}

\textbf{Comparison between KN and FINE:} By utilizing the model's unembedding parameters during computation, FINE effectively incorporates richer semantic representations. This integration enables FINE's localization results to exhibit stronger semantic alignment with the input context. To illustrate, consider the input example: \textit{MacApp, a product created by (Apple)}. As shown in Table \ref{KN&FINE}, FINE's localization outputs demonstrate more correlations with the ground truth. \textbf{Our results are aligned with} \citet{knowledgeneuron} \textbf{and} \citet{fine}. Additionally, an intriguing discovery is that both KN and FINE identify the neurons corresponding to the subject in the prompt. The results in Appendix \ref{knig} also support this finding. \textbf{The mutual corroboration seen in different methods further demonstrates the effectiveness of Know-MRI.}

We include the results of other interpretation methods in Appendix \ref{Additional_Results}. Generally, user-friendly UI-based usage allows users to comprehensively analyze the knowledge mechanisms of LLMs.

\subsection{Extended Application}
To further verify the potential utility of Know-MRI, we conduct capability localization experiments using Know-MRI. Specifically, code-based usage of Know-MRI is used in the experiments.

\paragraph{Experimental Setup} Our experiment involves the interpretation of Llama2-7B \citep{llama2} using the capability knowledge datasets (GSM8K and Emotion). The contribution of $j^{th}$ neuron $\omega^{l, j}$ at layer $l$ under the dataset $\mathcal{D}=\{(x=[x_1, \cdots, x_X], y=[y_1, \cdots, y_Y])\}$ is computed as:

\begin{small}
\begin{equation*}
\begin{split}
    &Score(\omega^{l, j}) =\\
     &\mathbb{E}_{(x, y)\in \mathcal{D}}\left[\frac{1}{Y}\frac{1}{S}\sum_{m=1}^{Y}\overline{\omega^{l, j}_{Z_m}[z_m]} \sum_{n=0}^{S}\frac{\partial P_{z, y_{m}}(\frac{n}{S} \overline{\omega^{l, j}_{Z_m}[z_m]})}{\partial \omega^{l, j}_{Z_m}[z_m]}\right], \\
    &z_m = x\oplus y_{0:m-1}
\end{split}
\label{eq_method}
\end{equation*}
\end{small}

where $x$ is the input prompt and $y$ is the corresponding ground truth. $\omega^{l, j}_{Z_m}[z_m]$ is the activation value of neuron $\omega^{l, j}$ and $\oplus$ means a splice of two text. Other settings are aligned with \citet{capability}. In the experiment, we employ the code-based usage methodology of Know-MRI. We use the overlap and IOU as location consistency ratio. Specifically, for two sets of neurons a, b located under different subset from the same dataset $\mathcal{D}$: 
\begin{equation*}
    overlap=\frac{\frac{|a\cap b|}{|a|}+\frac{|a\cap b|}{|b|}}{2}, 
IoU=\frac{|a\cap b|}{|a\cup b|}.
\end{equation*}
The location consistency ratio refers to the fidelity of a localization method to a dataset.
 \begin{figure}[!ht]
		\centering
    \begin{subfigure}{0.95\linewidth}
        \centering
        \includegraphics[width=\linewidth]{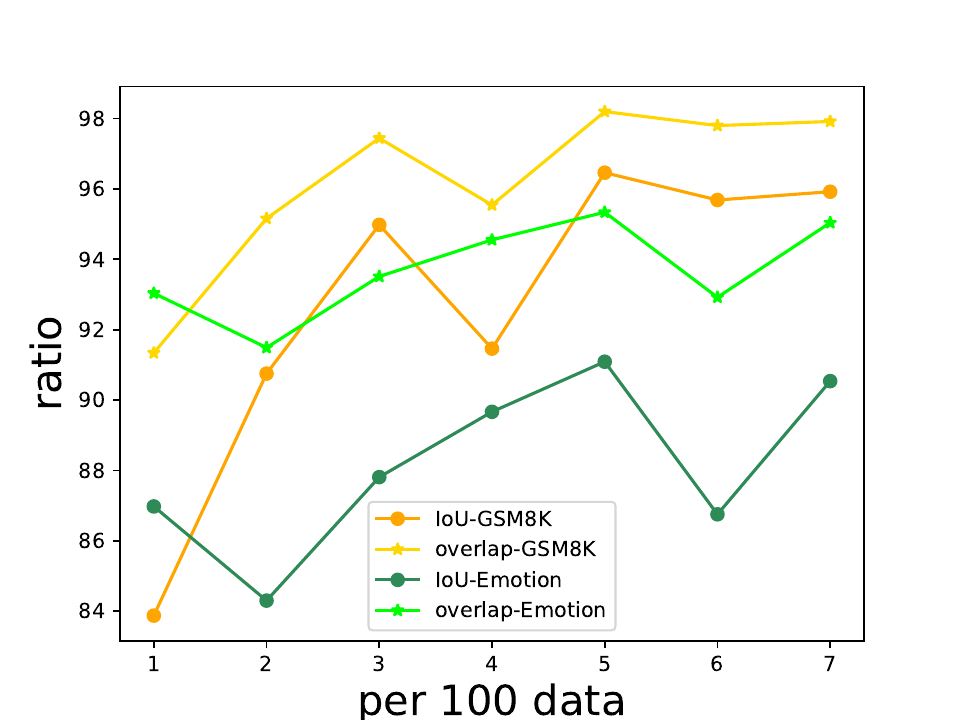}
    \end{subfigure}
		\caption{The relationship between location consistency ratio and the number of data.}
        \label{ourcapability}
\end{figure}
\paragraph{Result} Figure \ref{ourcapability} demonstrates that the location consistency ratio will gradually converge with increasing data. This result is the same as \citet{capability}. On the GSM8K dataset, the overlap and IOU scores are \textbf{98\%} and \textbf{96\%}, respectively. Meanwhile, on the Emotion dataset, these metrics reach \textbf{94\%} and \textbf{90\%}. We also provide the visualization of capability neurons in the Appendix \ref{Visualisation_Neurons}. Additionally, we conduct the neuron enhancement experiments in Table \ref{enhance}, which are similar with \citet{capability}. Specifically, we fine-tune the neurons whose contribution scores lie outside the range of 3 and 6 standard deviations $\sigma$. After 10 epochs, the located performance surpasses that of fine-tuning an equivalent quantity of random neurons and all the neurons excluding the localized ones (w/o located).  \textbf{Generally, the code-based usage of Know-MRI can effectively support users in customized experiments.}

\begin{table}[!ht]

	\begin{center}
        \resizebox{\linewidth}{!}{
		\begin{tabular}{cc|cccc}
			\toprule[2pt]\hline 
			\multirow{2}{*}{Model}&\multirow{2}{*}{Method}&\multicolumn{4}{c}{$epoch=10$} \\ &&GSM8K&Emotion&Code25K&$Avg.$ \\\hline
            \multirow{3}{*}{Llama2-7B ($\sigma=6$)}&random&
            5.25&14.99&\underline{53.05}&24.43 \\
            &w/o located&
            \underline{25.06}&\textbf{49.99}&46.48&\underline{40.51} \\
            &located&
            \textbf{25.56}&\underline{44.13}&\textbf{55.66}&\textbf{41.78} \\ \hline

            \multirow{3}{*}{Llama2-7B ($\sigma=3$)}&random&
            23.75&\underline{26.79}&\underline{53.47}&\underline{34.67} \\
            &w/o located&
            \underline{25.19}&19.29&42.77&29.08 \\
            &located&
            \textbf{26.31}&\textbf{51.63}&\textbf{56.02}&\textbf{44.65} \\ \hline

			\bottomrule[2pt]
		\end{tabular}
  }
        \caption{Enhancement experiment on different sets of neurons with 10 epochs. In the table, located neurons with different standard deviations $\sigma$,  equivalent random neurons and all the neurons excluding
the localized ones (w/o located) are enhanced. The best results are in \textbf{bold} and \underline{underline} means the suboptimal.}
        \label{enhance}
	\end{center}
 \end{table}

\subsection{Human Evaluation}
To comprehensively evaluate the effectiveness of Know-MRI, we invite ten independent researchers from the interpretation community who are not involved in this project. 

\paragraph{Experimental Setup} The researchers are allowed to use each toolkit freely. The evaluation framework consisted of four key dimensions: input diversity (ID), input flexibility (IF), method diversity (MD), and user-friendliness (UF). The max score is 5. The questionnaire can be found at our \href{https://docs.google.com/forms/d/e/1FAIpQLSepRhQXfVYklQHWUzu5IbQRLH0d8--BdNJHVK9SXzmDnUKOaA/viewform?usp=sharing\&ouid=103993125125082753123}{Google Forms}.

\begin{figure}[!ht]
	\centering
        \includegraphics[width=0.95\linewidth]{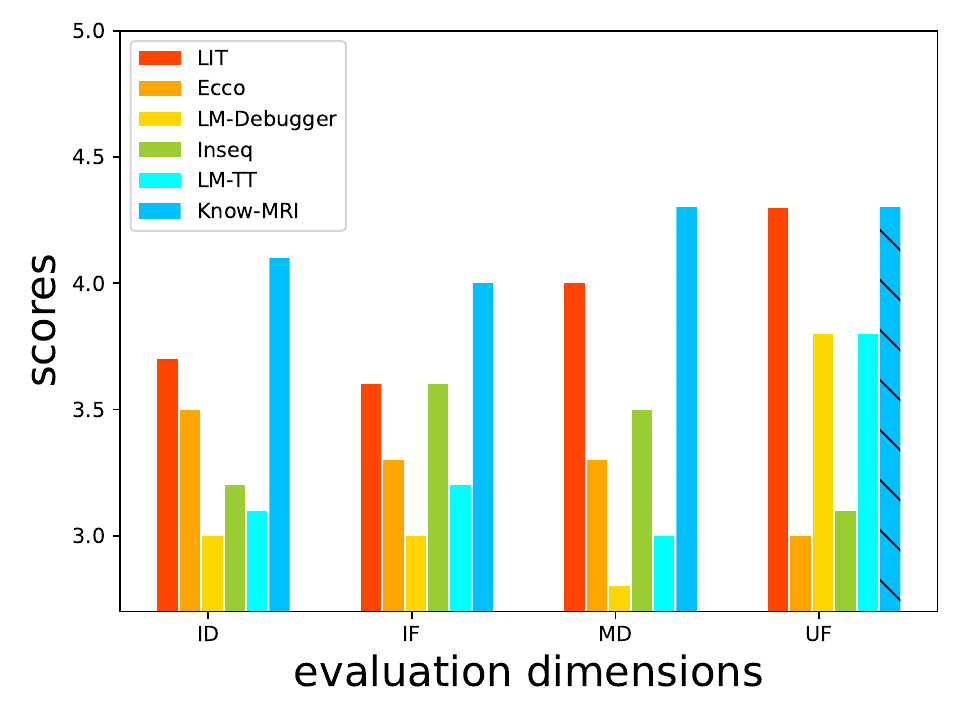}
		\caption{Human evaluation on existing toolkits.}
        \label{Human_score}
\end{figure}

\paragraph{Result} From Figure \ref{Human_score}, \textbf{results indicate that Know-MRI is highly evaluated in terms of user experience}.

\section{Conclusion}

Know-MRI is a comprehensive toolkit for analyzing knowledge mechanisms in LLMs. It is organized around three core components—models, datasets, and interpretation methods—with extensible interfaces for community development.  We also provide dual interaction modes: a UI-based interface and code-based usage. Case studies and human evaluations demonstrate Know-MRI's holistic design and usability advantages.


\section*{Acknowledgments}
This work was supported by the National Key R\&D Program of China (No. 2022ZD0160503) and Beijing Natural Science Foundation (L243006) and the National Natural Science Foundation of China (No. 62406321).

\bibliography{custom}
\newpage
\appendix

\section{Appendix / Interpretation Datasets} 
\label{datasetlisted}
To systematically investigate the knowledge mechanisms in LLMs, researchers have developed diverse datasets across multiple categories. The foundational datasets primarily focus on knowledge representation types, including: 1) commonsense knowledge \citep{zsre, pep3k, ROME, 20Q}, 2) biased knowledge \citep{biasneuron}, 3) counterfactual knowledge \citep{ROME}, 4) conceptual knowledge \citep{Conceptual-Knowledge}, etc. In addition, substantial efforts have been devoted to developing capability-oriented datasets for assessing specific LLM's capabilities, such as mathematical reasoning \citep{cobbe2021gsm8k, yu2023metamath}, sentiment understanding \citep{imdb, emotion}, and multilingual translation \citep{opus_books, opus_100}. 

\section{Appendix / Datasets Involved}
\label{dataset}
Here are datasets involved in Know-MRI:

\paragraph{ZsRE} ZsRE \citep{zsre} is prepared for zero-shot relation extraction task.

\paragraph{PEP3k} PEP3K \citep{pep3k} is a physical plausibility commonsense dataset with positive and negative labels.

\paragraph{Known-1000} Known-1000 \citep{ROME} includes a large amount of question pairs based on common sense, facts, and background knowledge, as well as the knowledge triples.

\paragraph{20Q} 20Q is a collection of 20 Questions style games, crowdsourced by expert.

\paragraph{Concept edit} Concept edit \citep{Conceptual-Knowledge} dataset is prepared for editing concept knowledge.

\paragraph{CounterFact} CounterFact \citep{ROME} dataset consists of counterfactual information based on Wikidata.

\paragraph{Bias neuron data} Bias neuron data \citep{biasneuron} contains bias quiz pairs to detect biased neurons in the LLM.

\paragraph{GSM8K} GSM8K \citep{cobbe2021gsm8k} contains approximately 8,000 elementary math problems with detailed solutions, designed to train mathematical reasoning models.

\paragraph{Meta Math} Meta Math \citep{yu2023metamath} focused on meta-learning for math problems, aimed at enhancing the model’s adaptive learning and reasoning capabilities.

\paragraph{Imdb} Imdb \citep{imdb} contains movie reviews and ratings, widely used for sentiment analysis and recommendation system research.

\paragraph{Emotion} Emotion \citep{emotion} with text data labeled with various emotions, suitable for sentiment analysis tasks, including social media posts and comments.

\paragraph{Opus Books} Opus Books \citep{opus_books} is a collection of copyright free books containing 16 languages.

\paragraph{Opus 100} Opus 100 \citep{opus_100} is an English-centric multilingual corpus covering 100 languages.

\section{Appendix / Template Keys}
\label{template_keys}

\begin{figure}[!ht]
		\centering
		\includegraphics[width=0.48\textwidth]{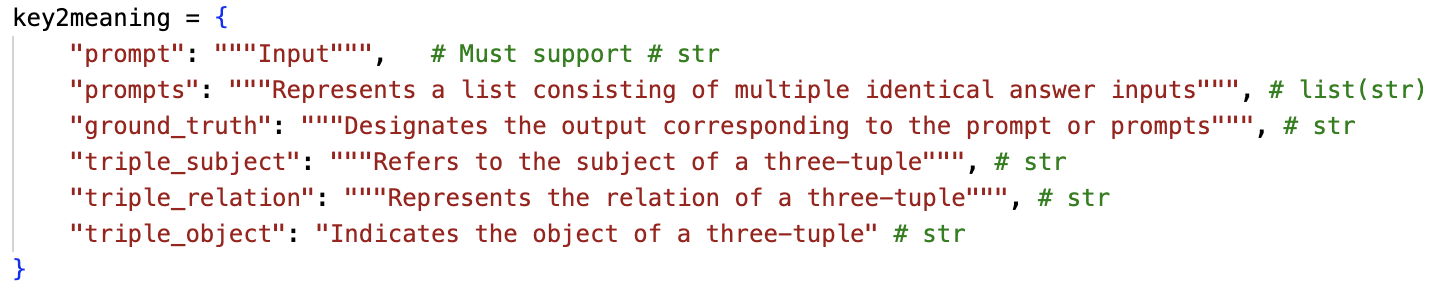}
		\caption{The supportive template keys and their meaning of Know-MRI. Users can also add corresponding keys as needed.}
        \label{demo_keys}
\end{figure}

Through extensive research on diverse datasets, we have identified several key inputs supported by existing interpretation methods. As demonstrated in Figure \ref{demo_keys}, these keys provide a foundational framework for dataset construction. Meanwhile, researchers are encouraged to extend this taxonomy by incorporating domain-specific parameters that align with their particular experimental requirements.

\section{Appendix / Additional Results on the Sample of Know 1000}
\label{Additional_Results}

\subsection{Comparison between Causal Tracing and Integrated Gradients}
\label{knig}
\begin{figure}[!ht]
		\centering
    \begin{subfigure}{0.9\linewidth}
        \centering
        \includegraphics[width=\linewidth]{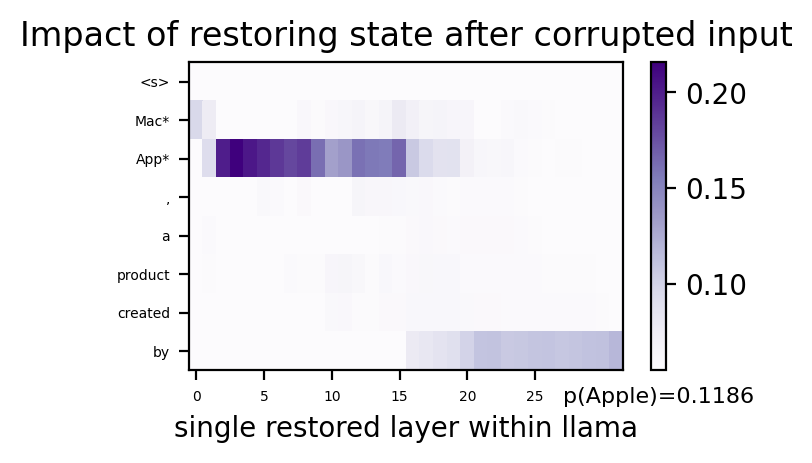}
        \caption{Impact of restoring state.}
    \end{subfigure}
    \begin{subfigure}{0.9\linewidth}
        \centering
        \includegraphics[width=\linewidth]{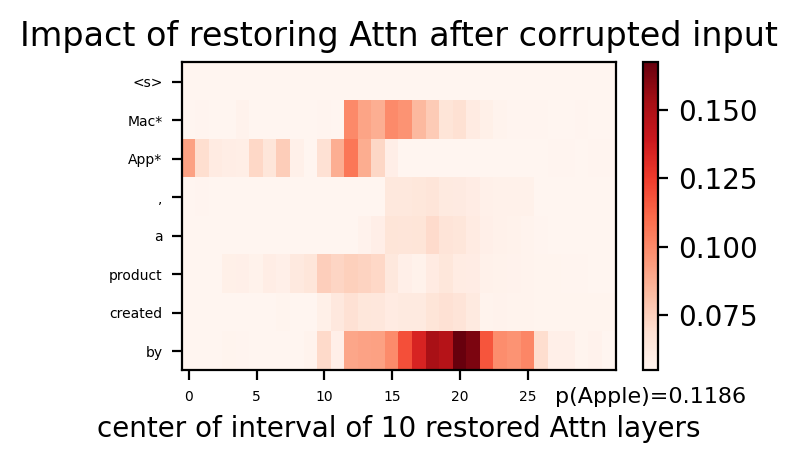}
        \caption{Impact of restoring attention layer.}
    \end{subfigure}
    \begin{subfigure}{0.9\linewidth}
        \centering
        \includegraphics[width=\linewidth]{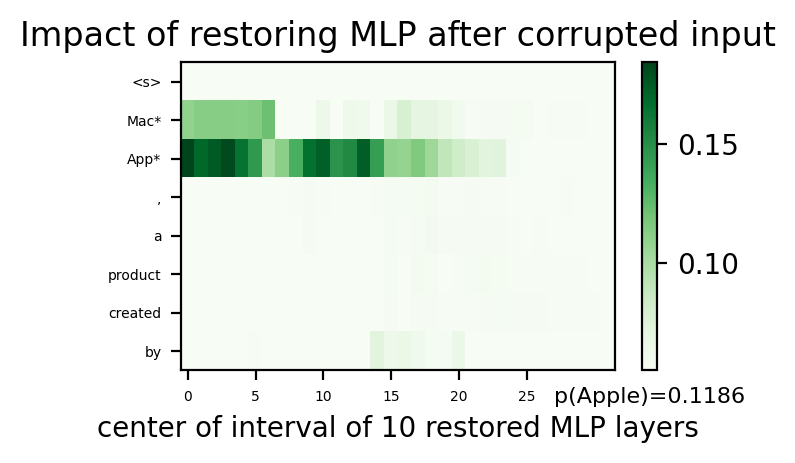}
        \caption{Impact of restoring MLP layer.}
    \end{subfigure}
		\caption{Causal Traceing's outputs.}
        \label{Causal Tracing}
\end{figure}

\begin{figure}[!ht]
		\centering
        \includegraphics[width=0.85\linewidth]{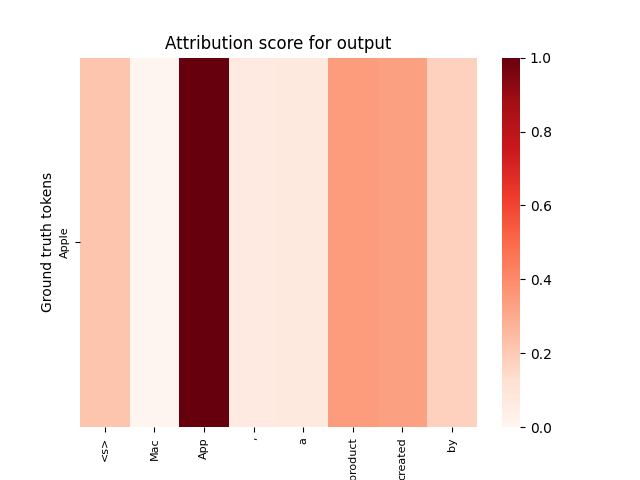}
        \caption{Attribution score computed by Integrated Gradients method.}
        \label{Integrated Gradients}
\end{figure}

Despite the differences in calculation methods, the results obtained by Causal Tracing \citep{ROME} and Integrated Gradients \citep{Integrated-Gradients} exhibit a certain degree of similarity. The results from Figure \ref{Causal Tracing} and Figure \ref{Integrated Gradients} collectively indicate: the impact of \textit{APP} token on the output is the most significant. Combining the results of neuron localization, we can find that for a factual input, the subject has a significant impact on the model's prediction.

From the Figure \ref{Causal Tracing}, the result of MLP demonstrates that the impact of the last subject token on the output is the most significant, \textbf{which also aligns with} \citet{ROME}. 


As shown in the figure \ref{Integrated Gradients}, the \textit{APP} token demonstrates the most significant influence on model outputs, which corroborates our conclusion from the previous section. \textbf{This alignment between experimental observation proves the effectiveness of Know-MRI.}

\subsection{Comparison between Logit Lens and PatchScopes}
Enabling LLMs to analyze their own hidden states via in-context learning, PatchScopes demonstrates the capability to predict the model's output at earlier layers. In the previously mentioned example, while Logit Lens requires processing through the final (\textbf{32nd}) layer to arrive at the prediction ``Apple'', PatchScopes successfully interprets hidden states as early as the \textbf{27th} layer to reach the same correct prediction. \textbf{This result is corresponding with} \citet{ghandeharioun2024patchscopes}.

\section{Appendix / Visualisation of Capacity Neurons}
\label{Visualisation_Neurons}
\begin{figure}[!ht]
		\centering
    \begin{subfigure}{0.9\linewidth}
        \centering
        \includegraphics[width=\linewidth]{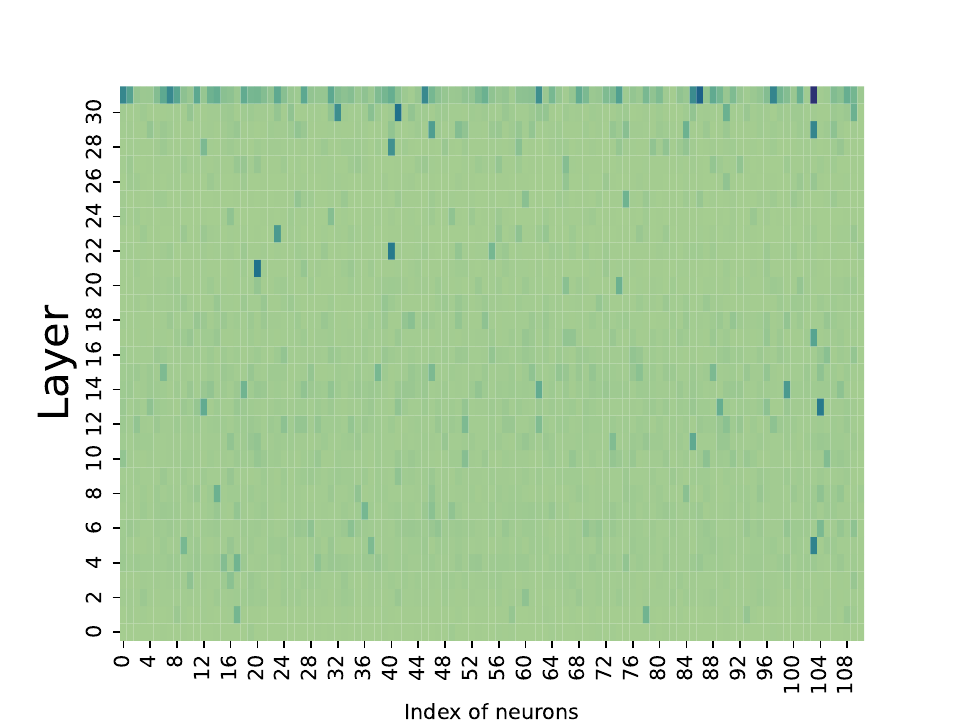}
        \caption{GSM8K}
    \end{subfigure}
    \begin{subfigure}{0.9\linewidth}
        \centering
        \includegraphics[width=\linewidth]{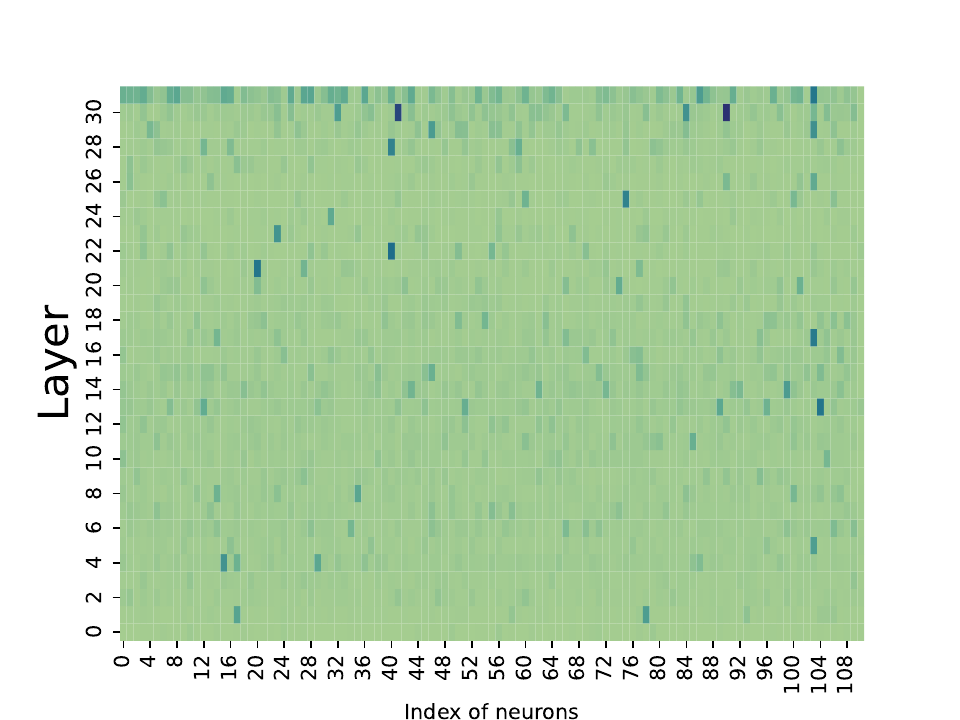}
        \caption{Emotion}
    \end{subfigure}
		\caption{We visualize the contribution score of the capacity neurons.}
\end{figure}

\end{document}